\documentclass{article} 

\usepackage[utf8]{inputenc} 

\usepackage{geometry} 
\usepackage{graphicx} 

\usepackage{booktabs} 
\usepackage{array} 
\usepackage{paralist} 
\usepackage{verbatim} 
\usepackage{subfig} 
\usepackage{siunitx} 
\usepackage{orcidlink} 
\usepackage{fancyhdr} 
\usepackage{sectsty}
\allsectionsfont{\sffamily\mdseries\upshape} 

\usepackage[nottoc,notlof,notlot]{tocbibind} 
\usepackage[titles,subfigure]{tocloft} 

\usepackage{amsthm} 
\usepackage{amssymb} 
\usepackage{amsmath} 
\usepackage[toc,page]{appendix} 
\usepackage{subfig} 
\usepackage[vlined, ruled]{algorithm2e} 
\usepackage{hyperref} 
\usepackage{lineno}

\DeclareMathOperator*{\argmax}{argmax} 
\DeclareMathOperator*{\argmin}{argmin} 
\newtheorem{theorem}{Theorem}[section]

\newtheorem{proposition}[theorem]{Proposition}

\begin{document}
\title{Estimating Gaussian Copulas with Missing Data}

\author{
Maximilian Kertel \orcidlink{0000-0003-3996-0642} \\ Battery Cell Competence Centre \\ BMW Group  \\ 80788 Munich, Germany \\ \and
Markus Pauly \orcidlink{0000-0002-0976-7190} \\ Institute for Mathematical Statistics and Industrial Applications \\ Faculty of Statistics \\ Technical University of Dortmund \\ 44221 Dortmund, Germany
}

\maketitle

\begin{abstract}
In this work we present a rigorous application of the Expectation Maximization algorithm to determine the marginal distributions and the dependence structure in a Gaussian copula model with missing data. We further show how to circumvent a priori assumptions on the marginals with semiparametric modelling. The joint distribution learned through this algorithm is considerably closer to the underlying distribution than existing methods. 
\end{abstract}

\section{Introduction}
\label{sec:Introduction}

\begin{figure}
\includegraphics[width=\textwidth]{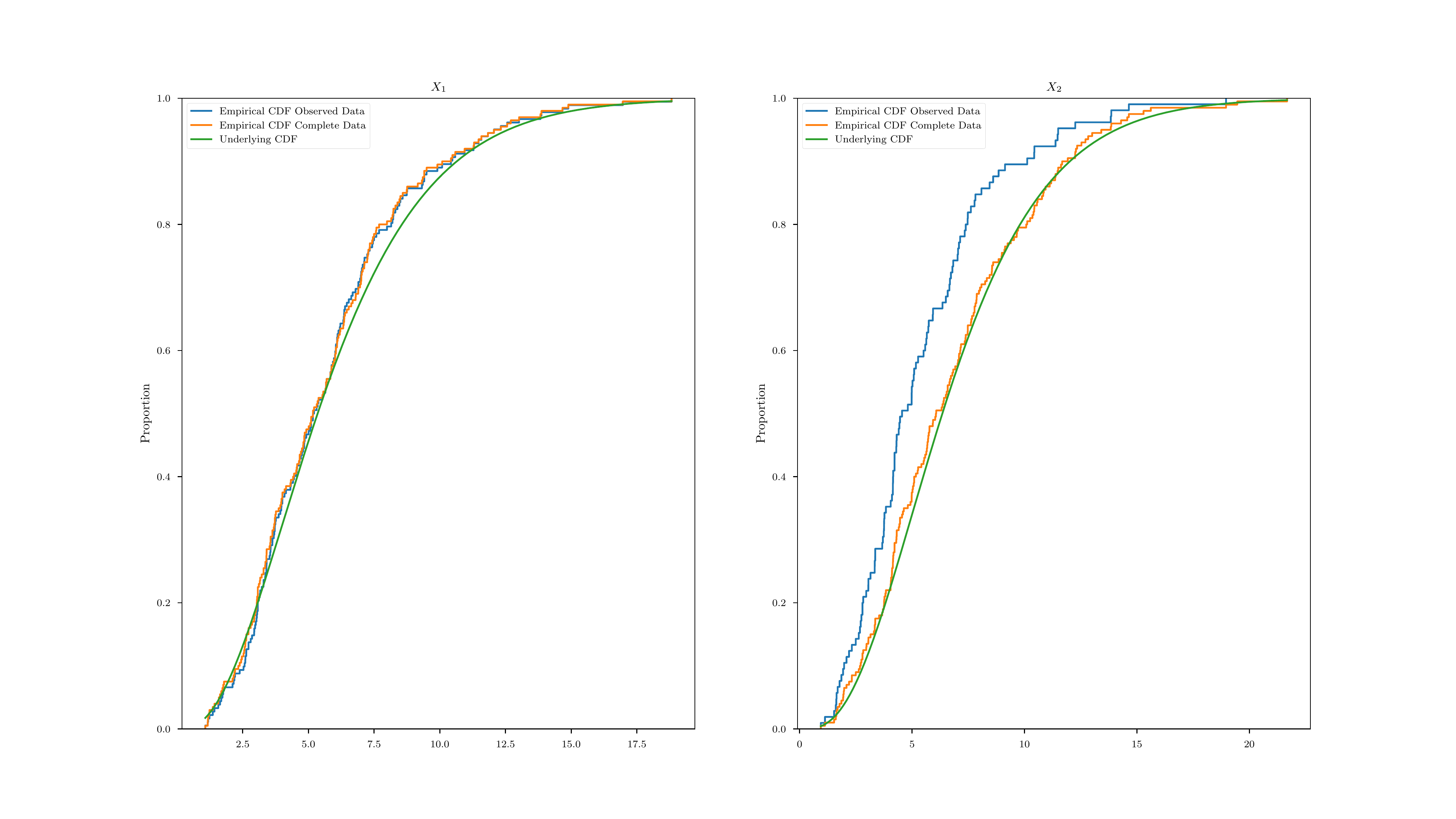}
\caption{Empirical cumulative distribution functions of observed (blue line) and complete data (orange line) and real, underlying distribution function (green line) for $X_1$ (left side) and $X_2$ (right side).}
\label{fig:introduction_example}
\end{figure}

Estimating the joint distribution with data Missing At Random (MAR) is a hard task. Usually, one applies strictly parametric methods, mostly relying on members of the exponential family such as the multivariate normal distribution. Its parameters can be determined by the Expectation Maximization (EM) algorithm (\cite{DempsterEM}). However, the misspecification error in the case of non-Gaussian data might be considerable. We can extend the normality assumption and assume a Gaussian copula model, liberating us from restrictions on the shape of the marginals. Figure \ref{fig:introduction_example} is based on such a distribution for the bivariate case. Here, the green lines show the underlying marginal cumulative distribution functions for the first (left) and the second component (right) of a two-dimensional random vector (X1, X2), respectively. We then generated $n=100$ observations from (X1, X2)  and calculated the corresponding empirical cumulative distribution functions (ecdf) (orange lines). To assess the influence of missings, we artificially chose some values to be Missing At Random (MAR) and recalculated the ecdfs based on the observed datapoints only (blue lines). In particular, the left column displaying $X_1$ is Missing Completely At Random (MCAR), while the missingness of the right column, which is displaying $X_2$, depends on $X_1$. \newline
\cite{SongEMCopula} propose an EM algorithm for this setting. However, their approach has two weaknesses. 
\begin{enumerate}
\item The presented algorithm simplifies by assuming that the marginals and the copula can be estimated separetely (compare Equation (6) in \cite{SongEMCopula} and Equation (\ref{CopulaEMAlgoEStep2}) in this paper) in the M-step.
\item As a parametrization of the marginals is necessary to apply the EM algorithm, they offer two options to model the marginals.
\begin{enumerate}
\item Fix the marginal distributions as the ecdf's of the \textbf{observed} data points and learn only the copula through the EM algorithm. This can lead to a heavily biased estimation, as the right-hand side of Figure \ref{fig:introduction_example} shows. The blue line depicting the ecdf of the observed data points of $X_2$ differs clearly from the underlying marginal cumulative distribution function of $X_2$, which is drawn in green.
\item Use parametric assumptions on the marginals. However, a Kolmogorov-Smirnov test with the correct underlying distribution of $X_2$ applied on the observed data points reveals a p-value of $7.559 * 10 ^{-5}$. Hence we would reject the (true) null hypothesis at every reasonable confidence level. Thus, this approach presumes a priori knowledge about the marginals, which rarely exists in practice.
\end{enumerate}
\end{enumerate}
It is the aim of the present paper to overcome these obstacles. Thereby, our contributions are two-fold.
\begin{enumerate}
\item We present a mathematically rigorous application of the EM algorithm in the Gaussian copula model. Similarly to complete data approaches, it estimates the marginals and the copula separately. However, the two steps are carried out repeatedly.
\item We propose a semiparametric approach for the marginal distributions. It allows us to learn their shape without possessing any a priori knowledge about them.
\end{enumerate}
The structure of the paper is as follows. In Section \ref{sec:CopulaModel} we review some background knowledge about copulas and the Gaussian copula in particular. We proceed by presenting the method (Section \ref{sec:EMAlgo}). In Section \ref{sec:SimulationStudy} we investigate its performance in a simulation study, where we also discuss the data generating process behind Figure \ref{fig:introduction_example}. Lastly, we conclude with a discussion and an outlook  in Section \ref{sec:Discussion}. All technical aspects and poofs of this paper are given in the Appendix \ref{sec:Appendix}.
\section{The Gaussian Copula Model}\label{sec:CopulaModel}
In the following we consider a $p$-dimensional data set $\{x_1, \ldots, x_N\} \subset \mathbb{R}^p$ of size $N$, where $x_l=\left(x_{l1}, \ldots, x_{lp}\right) \forall l = 1, \ldots, N$ and $x_1, \ldots, x_N$ are i.i.d. samples from a $p$-dimensional random vector $X = \left(X_1, \ldots, X_p\right)$ with joint distribution function $F$ and marginal distribution functions $F_1, \ldots, F_p$. The parameters of the marginals we denote as $\theta = \left(\theta_1, \ldots, \theta_p\right)$, where $\theta_j$ can be a vector itself and it is the parameter of $F_j$, so we write $F^{\theta_j}_j$. \newline 
For an observation $x$, we define $\mathbf{obs} \subset \{1, \ldots, p\}$ as the index set of the observed and $\mathbf{mis} \subset \{1, \ldots, p\}$ as the index set of the missing columns. Hence, $\mathbf{mis} \cup \mathbf{obs} = \{1, \ldots, p\}$ and $\mathbf{mis} \cap \mathbf{obs} = \emptyset$. $x_{l, \mathbf{obs}}$ and $x_{l, \mathbf{mis}}$ are then the subvectors containing the observable and unobservable values of the vector $x_l$. $R = \left(R_1, \ldots, R_p\right) \in \{0, 1\}^p$ is the random vector indicating if an entry is missing, where $R_i = 0$ if $X_i$ is missing. Further, we define $\phi$ to be the density function and $\Phi$ to be the distribution function of the one-dimensional standard normal distribution. $\Phi_{\mu, \Sigma}$ stands for the distribution function of a $p$-variate normal distribution with covariance $\Sigma \in \mathbb{R}^{p \times p}$ and mean $\mu \in \mathbb{R}^p$. To simplify the notation, we define $\Phi_{\Sigma} := \Phi_{0, \Sigma}$. For a matrix $A$, the entry of the $i$-th row and the $j$-th column, we denote as $A_{ij}$ and for index sets $\mathbf{S}, \mathbf{T} \subset \{1, \ldots, p\}$, $A_{\mathbf{S}, \mathbf{T}}$ is the submatrix of $A$ with row number in $\mathbf{S}$ and column number in $\mathbf{T}$. \newline
Throughout, we assume $F$ to be strictly increasing in every component and continuous. Therefore, for all $j \in \{1, \ldots, p\}$,  $F_j$ is strictly increasing and continuous and so is the existing inverse function $F_j^{-1}$.

\subsection{Properties}

Sklar's theorem \cite{Sklar} decomposes $F$ into its marginals $F_1, \ldots, F_p$ and its dependency structure $C$, by 
\begin{equation}
\label{eq:Sklar}
F(x_1, \ldots, x_p) = C\left(F_1(x_1), \ldots, F_p(x_p)\right).
\end{equation}
$C$ is a so-called copula, which means a $p$-dimensional distribution function with support $[0, 1]^p$, whose marginal distributions are uniform. Hence, every distribution belongs to an equivalence class, which contains all the distributions with the identical copula. In this paper we focus on the so-called Gaussian copulas, where 
\begin{equation}
\label{eq:GaussianCopula}
C_{\Sigma}(u_1, \ldots, u_p) = \Phi_{\Sigma}\left(\Phi^{-1}(u_1), \ldots, \Phi^{-1}(u_p)\right), \Sigma_{jj} = 1 \text{ }\forall j \in \{1, \ldots, p\}.
\end{equation}
and $\Sigma$ is of full rank. We can see that every equivalence class contains exactly one multivariate normal random vector $Y$ with $\mathbb{E}(Y_j) = 0$ and $\mathbb{E}(Y_j^2) = 1$ for all $j=1, \ldots, p$. Beyond all multivariate normal distributions that share the same correlation structure, every equivalence class contains other distributions that do not have Gaussian marginals. Hence, the Gaussian copula model provides us with an extension of the normality assumption.
Let's take a random vector $X$ from such an equivalence class. Under the transformation 
$$
Z := \Phi^{-1} \circ F\left(X\right) := \left(\Phi^{-1} \circ F_1 \left(X_1\right), \ldots, \Phi^{-1} \circ F_p\left(X_p\right)\right)
$$ 
it holds
\begin{equation}
\label{eq:GaussianCopulaTransformed}
\begin{aligned}
F_Z(z_1 \ldots, z_p) 
&= \mathbb{P}\left(Z_1 \leq z_1, \ldots, Z_p \leq z_p\right) \\ 
&= \mathbb{P}\left(X_1 \leq F_1^{-1}\left(\Phi\left(z_1\right)\right), \ldots, X_p \leq F_p^{-1}\left(\Phi\left(z_p\right)\right)\right) \\
&=F_X\left(F^{-1}_1\left(\Phi(z_1)\right), \ldots, F^{-1}_p\left(\Phi(z_p)\right)\right) \\
&= \Phi_\Sigma\left(\Phi^{-1}\left(F_1\left(F^{-1}_1\left(\Phi(z_1)\right)\right)\right), \ldots, \Phi^{-1}\left(F_p\left(F^{-1}_p\left(\Phi(z_p)\right)\right)\right)\right) \\
&= \Phi_\Sigma(z_1, \ldots, z_p)
\end{aligned}
\end{equation}
and hence $Z = \Phi^{-1} \circ F(X)$ is normally distributed with mean $0$ and covariance $\Sigma$. The two-step approaches given in \cite{GenestSemiparametricEstimation} and \cite{liu2012high} use this property and apply the following scheme:
\begin{enumerate}
\item Find consistent estimates $\hat{F_1}, \ldots, \hat{F_p}$ for the marginal distributions $F_1, \ldots, F_p$.
\item Find $\Sigma$ by estimating the covariance of the random vector $$Z = \left(\Phi^{-1}\left(\hat{F_1}\left(X_1\right)\right), \ldots, \left(\Phi^{-1}\left(\hat{F_p}\left(X_p\right)\right)\right)\right).$$
\end{enumerate}
From now on assume that the marginals of $X$ have existing density functions $f_1, \ldots, f_p$. Then, using Equation (\ref{eq:GaussianCopulaTransformed}) and a change of variables, we can derive the joint density function (see \cite{SongGaussianCopulaBasics})
\begin{equation}
\label{eq:DensityGaussianCopula}
f_{F_1, \ldots, F_p, \Sigma}(x_1, \ldots, x_p) = f(x_1, \ldots, x_p)  = |\Sigma|^{-\frac{1}{2}} \exp\left(-\frac{1}{2} z^T \left(\Sigma^{-1} - I\right) z \right) \prod_{j=1}^p f_j(x_j),
\end{equation}
where $z := \left(\Phi^{-1}\left(F_1(x_1)\right), \ldots, \Phi^{-1}\left(F_p(x_p)\right)\right)$. As in the case of the multivariate normal distribution, we can read off conditional independencies (\cite{liu2012high}) from the inverse of the covariance matrix $K := \Sigma^{-1}$, using the property
\begin{equation}
\label{eq:ConditionalIndependence}
K_{jk} = K_{kj} = 0 \iff X_j \perp X_k | \left\{X_i : i \in \{1, \ldots, p\} \setminus \{j, k\}\right\}.
\end{equation}
$K$ is called the precision matrix. In order to slim down the notation, we define for a subset $\mathbf{S} = \{s_1, \ldots, s_k\} \subset \{1, \ldots, p\}$ 
$$F_\mathbf{S}(x) := \left(F_{s_1}(x_{s_1}), \ldots, F_{s_k}(x_{s_k})\right)$$ and similarly 
$$\Phi^{-1}\left(F_\mathbf{S}(x)\right) := \left(\Phi^{-1}\left(F_{s_1}(x_{s_1})\right), \ldots, \Phi^{-1}\left(F_{s_k}(x_{s_k})\right)\right).$$
The conditional density functions have a closed form. 
\begin{proposition}[Conditional Distribution of Gaussian copula]\label{ConditionalDistributionGaussianCopula} Let $\mathbf{S} = \{s_1, \ldots, s_k\}$ and $\mathbf{T} = \{t_1, \ldots, t_{k'}\}$ be such that $\mathbf{T} \dot{\cup} \mathbf{S} = \{1, \ldots, p\}$.
\begin{enumerate}
\item The conditional density of $X_{\mathbf{T}} | X_{\mathbf{S}} = x_{\mathbf{S}}$ is given by 
$$f(x_\mathbf{T} | X_\mathbf{S} = x_\mathbf{S}) =  |\Sigma'|^{-\frac{1}{2}} \exp\left(-\frac{1}{2} (z_{\mathbf{T}} - \mu)^T \Sigma'^{-1} (z_{\mathbf{T}} - \mu)\right) \exp\left(\frac{1}{2} z_{\mathbf{T}}^T z_{\mathbf{T}}\right) \prod_{j \in \mathbf{T}} f_j(x_j), $$
where $\mu = \Sigma_{\mathbf{T}, \mathbf{S}}\Sigma_{\mathbf{S}, \mathbf{S}}^{-1}z_\mathbf{S}$, $\Sigma' = \Sigma_{\mathbf{T}, \mathbf{T}} - \Sigma_{\mathbf{T}, \mathbf{S}} \Sigma_{\mathbf{S}, \mathbf{S}}^{-1}\Sigma_{\mathbf{S}, \mathbf{T}}$, $z_\mathbf{T} = \Phi^{-1}\left(F(x_\mathbf{T})\right)$ and $z_\mathbf{S} = \Phi^{-1}\left(F(x_\mathbf{S})\right).$
\item $\Phi^{-1}\left(F_\mathbf{T}(X_\mathbf{T})\right) | X_\mathbf{S} = x_\mathbf{s}$ is normally distributed with mean $\mu$ and covariance $\Sigma'$.
\item The expectation of $h\left(X_\mathbf{T}\right)$ with respect to the density $f(x_\mathbf{T} | X_\mathbf{S} = x_\mathbf{S})$ can be expressed by
\begin{equation*}
\int h(x_{\mathbf{T}}) f(x_\mathbf{T} | X_\mathbf{S} = x_\mathbf{S}) dx_\mathbf{T}= \int h\left(F^{-1}\left(\Phi\left(z_\mathbf{T}\right)\right)\right)\phi_{\mu, \Sigma'}\left(z_\mathbf{T}\right)dz_\mathbf{T}.
\end{equation*}
\end{enumerate}
\end{proposition}
\noindent Using Proposition \ref{ConditionalDistributionGaussianCopula} it can be seen that the conditional distribution's copula is Gaussian as well. More importantly, we can derive an algorithm for sampling data points from the conditional distribution:

\begin{algorithm}[H]
\label{algo:SamplingGaussianCopula}
\SetAlgoLined
\KwIn{$x_\mathbf{S}, \Sigma, F_1, \ldots, F_p$}
\KwResult{$m$ samples of $X_\mathbf{T} | X_\mathbf{S} = x_\mathbf{S}$} 
Calculate $z_\mathbf{S} := \Phi^{-1}\left(F_\mathbf{S}(x_\mathbf{S})\right)$ \;
Calculate $\mu$ and $\Sigma'$ as in Proposition \ref{ConditionalDistributionGaussianCopula} using $z_\mathbf{S}, \Sigma, F_1, \ldots, F_p$  \;
Draw samples $\{z^1, \ldots, z^m\}$ from $\mathcal{N}(\mu, \Sigma')$ \;
\KwRet{\{$F_\mathbf{T}^{-1}\left(\Phi(z^1)\right), \ldots, F_\mathbf{T}^{-1}\left(\Phi(z^m)\right)\}$}
\caption{Sample from the conditional distribution of a Gaussian copula}
\end{algorithm}

The very last step follows by Proposition \ref{ConditionalDistributionGaussianCopula}, as it holds for any measurable $A \subset \mathbb{R}^{t_{k'}}$, that:
\begin{align*}
\mathbb{P}\left(X_\mathbf{T} \in A  | X_\mathbf{S} = x_\mathbf{S}\right) &= \int 1_{A}(x_\mathbf{T})  f(x_\mathbf{T} | X_\mathbf{S} = x_\mathbf{S}) dx_\mathbf{T} = \int 1_A\left(F^{-1}\left(\Phi\left(z_\mathbf{T}\right)\right)\right)\phi_{\mu, \Sigma'}\left(z_\mathbf{T}\right)dz_\mathbf{T}
\end{align*}


\section{The EM Algorithm in the Gaussian Copula Model}\label{sec:EMAlgo}
\subsection{The EM Algorithm}
Let $y_1, \ldots, y_N$ be an incomplete data set following a parametric distribution with parameter $\psi$ and corresponding density function $g_\psi(\cdot)$, where observations are MAR. The EM algorithm \cite{DempsterEM} finds a local optimum of the log-likelihood function
\begin{align*}
\sum_{l=1}^N\ln\left(g_\psi\left(y_{l, \mathbf{obs}}\right)\right) &= \sum_{l=1}^N \int \ln\left(g_\psi\left(\left(y_{l, \mathbf{obs}}, y_{l, \mathbf{mis}}\right)\right)\right) g_\psi\left(y_{l, \mathbf{mis}} | Y_{l, \mathbf{obs}} = y_{l, \mathbf{obs}}\right) dy_{l, \mathbf{mis}}  \\ 
&= \sum_{l=1}^N \mathbb{E}_{\psi}\left(\ln\left(g_\psi\left(\left(y_{l, \mathbf{obs}}, y_{l, \mathbf{mis}}\right)\right)\right)| Y_{l, \mathbf{obs}} = y_{l, \mathbf{obs}}\right).
\end{align*}
After choosing a start value $\psi^0$, it does so by iterating the following two steps.
\begin{enumerate}
\item E-Step: Calculate 
\begin{equation}
\label{eq:EMGeneralEStep}
\lambda(\psi|y_1, \ldots, y_p, \psi^t) := \sum_{l=1}^N\mathbb{E}_{\psi^{t}}\left(\ln\left(g_\psi\left(\left(y_{l, \mathbf{obs}}, y_{l, \mathbf{mis}}\right)\right)\right) | Y_{l, \mathbf{obs}} = y_{l, \mathbf{obs}}\right).
\end{equation}
\item M-Step: Set 
\begin{equation}
\label{eq:EMGeneralMStep}
\psi^{t+1} = \argmax_{\psi}\lambda(\psi|y_1, \ldots, y_p, \psi^t)
\end{equation}
and $t = t +1$.
\end{enumerate}
For our purpose, there exist two extensions of interest.
\begin{itemize}
\item In many cases there is no closed formula for the right-hand side of Equation (\ref{eq:EMGeneralEStep}). If it is possible to sample from the conditional distribution, then one can use Monte Carlo integration \cite{WeiStochasticEM} as an approximation, which is called the Monte Carlo EM algorithm.
\item If $\psi = \left(\psi_1, \ldots, \psi_v\right)$ and the joint maximization of (\ref{eq:EMGeneralMStep}) with respect to $\psi$ is not feasible, \cite{MengECM} proposed a sequential maximization. In that case, we maximize (\ref{eq:EMGeneralMStep}) with respect to $\psi_i$ holding $\psi_1 = \psi_1^{t+1}, \ldots, \psi_{i-1} = \psi_{i-1}^{t+1}, \psi_{i+1} = \psi_{i+1}^{t}, \ldots, \psi_{v} = \psi_{v}^{t}$ fixed, before continuing with $\psi_{i + 1}$. This is called the Expectation Conditional Maximization (ECM) algorithm.
\end{itemize}


\subsection{Applying the ECM-Algorithm on the Gaussian Copula Model}
\label{EMAlgoCopulaModel}
As we need a full parametrization of the Gaussian copula model for the EM algorithm, we choose parametric marginal distributions $F_1^{\theta_1}, \ldots, F_p^{\theta_p}$ with densities $f_1^{\theta_1}, \ldots, f_p^{\theta_p}$. According to Equation (\ref{eq:DensityGaussianCopula}), the joint density with respect to the parameters $\theta= \left(\theta_1, \ldots, \theta_p\right)$ and $\Sigma$ has the form
\begin{equation}
\label{eq:DensityGaussianCopulaParametric}
f_{\theta, \Sigma}(x_1, \ldots, x_p) = |\Sigma|^{-\frac{1}{2}} \exp\left(-\frac{1}{2} z_\theta^T \left(\Sigma^{-1} - I\right) z_\theta \right) \prod_{j=1}^p f^{\theta_j}_j(x_j),
\end{equation}
where $z_\theta := \left(\Phi^{-1}\left(F^{\theta_1}_1\left(x_1\right)\right), \ldots, \Phi^{-1}\left(F^{\theta_p}_p\left(x_p\right)\right)\right)$. Section \ref{sec:SemiparametricMarginals} will describe how we can keep the flexibility for the marginals despite the parametrization. But first we outline the EM algorithm for general parametric marginal distributions.
\subsubsection{E-Step}
In the following, define $K:= \Sigma^{-1}$ and $K^{t} := \Sigma^{t^{-1}}$. For simplicity, we focus on one particular observation $x$. According to Equation (\ref{eq:EMGeneralEStep}) and (\ref{eq:DensityGaussianCopulaParametric}), it holds (with $\psi = \left(\theta, \Sigma\right)$ and $x$ taking the role of $\left(y_1, \ldots, y_p\right)$)
\begin{equation}
\label{CopulaEMAlgoEStep}
\begin{split}
\lambda(\theta, \Sigma | \theta^t, \Sigma^t, x) &= \mathbb{E}_{\theta^t, \Sigma^t}\left(\ln\left(f_{\theta, \Sigma}\left(\left(x_\mathbf{obs}, x_\mathbf{mis}\right)\right)\right)| X_\mathbf{obs} = x_\mathbf{obs}\right) \\ 
&= - \frac{1}{2} \ln\left(|\Sigma|\right) \\ 
& \quad -\frac{1}{2} \mathbb{E}_{\Sigma^t, \theta^t } \left(z_{\theta}^T  \left(K - I\right)  z_\theta | X_\mathbf{obs} = x_\mathbf{obs} \right) \\
& \quad + \sum_{j=1}^p \mathbb{E}_{\Sigma^t, \theta^t}\left(\ln\left(f^{\theta_j}_j(x_j)\right) | X_\mathbf{obs} = x_\mathbf{obs}\right).
\end{split}
\end{equation}
The first and the last summand depend only on $\Sigma$ and $\theta$, respectively. Thus, of special interest is the second summand, for which we observe by Proposition \ref{ConditionalDistributionGaussianCopula}
\begin{equation}
\label{CopulaEMAlgoEStep2}
\mathbb{E}_{\Sigma^t, \theta^t } \left(z_\theta^T  \left(K - I\right)  z_\theta | X_\textbf{obs} = x_\textbf{obs} \right) =  \int z_{\theta, \theta^t}^T  \left(K - I\right)  z_{\theta, \theta^t}  \phi_{\mu, {\Sigma^t}'}\left(q_\textbf{mis}\right) dq_\textbf{mis},
\end{equation}
where $$ z_{\theta, \theta^t}:= \left(\Phi^{-1}\left(F^{\theta_1}_1\left(F^{{\theta^t_1}^{-1}}_1\left(\Phi(q_1)\right)\right)\right), \ldots, \Phi^{-1}\left(F^{\theta_p}_p\left(F^{{\theta^t_p}^{-1}}_p\left(\Phi(q_p)\right)\right)\right)\right),$$
$\mu = \Sigma_{\textbf{mis},\textbf{obs}}\Sigma_{\textbf{obs},\textbf{obs}}^{-1}\Phi^{-1}\left({F_\textbf{obs}^{\theta^t}} \left(x_\textbf{obs}\right)\right)$ and ${\Sigma^t}' = \Sigma^t_{\textbf{mis}, \textbf{mis}} - \Sigma^t_{\textbf{mis}, \textbf{obs}} \left({\Sigma^t_{\textbf{obs},\textbf{obs}}}\right)^{-1}\Sigma^t_{\textbf{obs}, \textbf{mis}}$. At this point \cite{SongEMCopula} neglect, that in general 
$$F^{\theta^t_k} \neq F^{\theta_k}, k = 1, \ldots, p$$
holds and hence (\ref{CopulaEMAlgoEStep2}) is depending not only on $\Sigma$ but also on $\theta$. This has let us reconsider their approach as we describe below.
\subsubsection{M-Step}
We have encountered, that the joint optimization with respect to $\theta$ and $\Sigma$ is difficult, as there is no closed-form solution for (\ref{CopulaEMAlgoEStep}). We circumvent this problem by sequentially optimizing with respect to $\Sigma$ and $\theta$, applying the ECM algorithm. The maximization routine is the following.
\begin{enumerate}
\item Set $\Sigma^{t+1} = \argmax_{\Sigma} \sum_{l=1}^N\lambda(\theta^t, \Sigma | \theta^t, \Sigma^t, x_l)$.
\item Set $\theta^{t+1} = \argmax_{\theta}\sum_{l=1}^N \lambda(\theta, \Sigma^{t+1} | \theta^t, \Sigma^t, x_l)$. 
\end{enumerate}
The reader might notice, that this is a two-step approach consisting of first estimating the copula encoded in $\Sigma$ and then estimating the marginals by finding $\theta$. However, both steps are executed iteratively, which is typical for the EM algorithm. 


\subsubsection*{Estimating $\Sigma$}
As we are maximizing Equation (\ref{CopulaEMAlgoEStep}) with respect to $\Sigma$ with a fixed $\theta = \theta^t$, the last summand can be neglected. By a change of variables argument we show in Proposition \ref{ConditionalExpectationGaussianizedEM}, that 
$$- \frac{1}{2} \ln\left(|\Sigma|\right) -\frac{1}{2} \mathbb{E}_{\Sigma^t, \theta^t } \left({z_{\theta^t}}^T  \left(K - I\right)  z_{\theta^t} | X_\mathbf{obs} = x_\mathbf{obs} \right) = -\frac{1}{2}\ln(|\Sigma|) - \frac{1}{2}tr\left(\Sigma^{-1}V\right),$$
where $V$ depends on $\Sigma^t$ and $z_{\theta^t, {\mathbf{obs}}} = \Phi^{-1}\left(F^{\theta^t}\left(x_{\mathbf{obs}}\right)\right)$. Considering all observations, we search for 
\begin{equation}
\label{eq:SolutionSigma}
\begin{aligned}
\Sigma^{t+1} &= \argmax_{\Sigma, \Sigma_{ll} = 1 \forall l=1, \ldots, p} \frac{1}{N}\sum_{l=1}^N-\frac{1}{2}\ln(|\Sigma|) - \frac{1}{2}tr\left(\Sigma^{-1}V_l\right) \\
&= \argmax_{\Sigma, \Sigma_{ll} = 1 \forall l=1, \ldots, p} -\frac{1}{2}\ln(|\Sigma|) - \frac{1}{2}tr\left(\Sigma^{-1} \frac{1}{N}\sum_{l=1}^N V_l\right)
\end{aligned}
\end{equation}
with $V_l$ as in Proposition \ref{ConditionalExpectationGaussianizedEM}, but for observation $x_l$. The maximizer depends on the statistic $S := \frac{1}{N}\sum_{l=1}^N V_l$ only. Generally, the maximization can be formalized as a convex optimization problem, which can be solved by a gradient descent. However, the properties of this estimator are not understood (for example a scaling of $S$ by $a \in \mathbb{R}_{>0}$ leads to a different solution, see Appendix \ref{sec:appendix_max_sigma}). To overcome this issue, we instead approximate the solution by the correlation matrix 
$$ \argmax_{\Sigma, \Sigma_{ll} = 1 \forall l=1, \ldots, p} -\frac{1}{2}\ln(|\Sigma|) - \frac{1}{2}tr\left(\Sigma^{-1} S \right) \approx P S P,$$ 
where $P \in \mathbb{R}^p$ is the diagonal matrix with entries $P_{jj} = \frac{1}{\sqrt{S_{jj}}}, \forall j = 1, \ldots, p$. This was also proposed in \cite[Section~2.2]{GuoCorrelation}.

\subsubsection*{Maximizing with respect to $\theta$}
We now focus on finding $\theta^{t+1}$, which maximizes
$$\sum_{l=1}^N \lambda(\theta, \Sigma^{t+1} | \theta^t, \Sigma^t, x_l) = \sum_{l=1}^N\mathbb{E}_{\theta^t, \Sigma^t}\left(\ln\left(f_{\theta, \Sigma^{t+1}}\left(\left(x_{l,\mathbf{obs}}, x_{l, \mathbf{mis}}\right)\right)\right)| X_{l, \mathbf{obs}} = x_{l,\mathbf{obs}}\right)
$$
with respect to $\theta$. As there is in general no closed form for the expectations, we approximate them with Monte Carlo integration. Again, we start by considering a single observation $x$ to simplify the terms. Employing Algorithm \ref{algo:SamplingGaussianCopula}, we receive $m$ samples $x_{\textbf{mis}}^1, \ldots, x_{\textbf{mis}}^m$ from the distribution of $X_{\mathbf{mis}} | X_{\mathbf{obs}} = x_{\mathbf{obs}}$ under the parameters $\theta^t$ and $\Sigma^t$. We set $x^k_{\mathbf{obs}} = x_{\mathbf{obs}} \text{ } \forall k = 1, \ldots, m$. Then, 
\begin{equation}\label{CopulaEMAlgoMonteCarlo}
\begin{aligned}
\lambda(\theta, \Sigma^{t+1} | \theta^t, \Sigma^t, x)  \approx \frac{1}{M} \sum_{m=1}^M &-\frac{1}{2}\left(\Phi^{-1}\left(F^{\theta_1}_1(x^m_1)\right), \ldots, \Phi^{-1}\left(F^{\theta_p}_p(x^m_p)\right)\right)^T \\
&\qquad\left(K^{t+1} - I\right) \\ 
&\qquad\left(\Phi^{-1}\left(F^{\theta_1}_1(x^m_1)\right), \ldots, \Phi^{-1}\left(F^{\theta_p}_p(x^m_p)\right)\right) \\ 
&+ \sum_{j=1}^p \ln\left(f^{\theta_j}_j(x^m_j)\right).
\end{aligned}
\end{equation}
Hence, considering all observations, we set
\begin{equation}
\label{CopulaEMAlgoMonteCarloSum}
\begin{aligned}
\theta^{t+1} = \argmax_\theta \frac{1}{M} \sum_{l=1}^N\sum_{m=1}^M  &-\frac{1}{2}\left(\Phi^{-1}\left(F^{\theta_1}_1(x^m_{l1})\right), \ldots, \Phi^{-1}\left(F^{\theta_p}_p(x^m_{lp})\right)\right)^T \\
& \qquad \left(K^{t+1} - I\right) \\ 
& \qquad \left(\Phi^{-1}\left(F^{\theta_1}_1(x^m_{l1})\right), \ldots, \Phi^{-1}\left(F^{\theta_p}_p(x^m_{lp})\right)\right)  \\
&+ \sum_{j=1}^p \ln\left(f^{\theta_j}_j(x^m_{lj})\right). 
\end{aligned}
\end{equation}
We emphasize, that not only sampling but also the evaluation and the calculation of derivatives of the right hand side of Equation (\ref{CopulaEMAlgoMonteCarloSum}) can be parallelized. \newline 
The idea of using Monte Carlo integration is similar to the Monte Carlo EM algorithm introduced by \cite{WeiStochasticEM}. However, note that we only use the Monte Carlo samples to update the parameters of the marginal distributions $\theta$.
\newline
We would also like to point out some interesting aspects about Equations (\ref{CopulaEMAlgoMonteCarlo}) and (\ref{CopulaEMAlgoMonteCarloSum}):
\begin{itemize}
\item The summand $\sum_{j=1}^p \ln\left(f^{\theta_j}_j(x^m_{lj})\right)$ describes how well the marginal distributions fit to the (one-dimensional) data.
\item The first summand adjusts for the dependence structure in the data. If all observations at step $t+1$ are assumed to be independent, then $K^{t + 1} = I$ and this term is $0$. 
\item More generally, the derivative $\frac{\partial \lambda(\theta, \Sigma^{t+1} | \theta^t, \Sigma^t, x)}{\partial\theta_j}$ depends on $\theta_k$ if and only if $K^{t+1}_{jk} \neq 0$. That means, if $\Sigma^{t+1}$ implies conditional independence of column $j$ and $k$ given all other columns (Equation (\ref{eq:ConditionalIndependence})), the optimal $\theta_j$ can be found without considering $\theta_k$. 
\end{itemize}
\subsection{Modelling with Semiparametric Marginals}
\label{sec:SemiparametricMarginals}
The algorithm of Section \ref{EMAlgoCopulaModel} depends on a parametrization of the marginals. We have seen in Section \ref{sec:Introduction}, that we cannot deduce the parametric family of the marginal distributions by the observed data points. If there is a priori knowledge about the parametrization of the marginals, then we can plug it into the formulae of Section \ref{EMAlgoCopulaModel}. However, as this prior knowledge for the marginals rarely exists (they might not even belong to a parametric family), we propose the usage of semiparametric mixture models.
In particular, we are using a parametrization of the form
\begin{equation}
\label{eq:MixtureMarginalsSimplified}
F_j^{\theta_j}(x_j) = \frac{1}{g} \sum_{k=1}^g \Phi\left(\frac{x_j - \theta_{jk}}{\sigma_{j}}\right), \theta_{j1} \leq \ldots \leq \theta_{jg}, \forall j = 1, \ldots, p,
\end{equation}
where $\sigma_j$ is a hyperparameter and the ordering of the $\theta_{ji}$ ensures the identifiability of the distribution (\cite{MclachlanFiniteMixtures}). \newline
Using mixture models for density estimation is not new (e.g. \cite{MclachlanFiniteMixtures}, \cite{HwangNonparametric}, \cite{ScottMultidimensionalDensityEstimation}). As \cite{ScottMultidimensionalDensityEstimation} note, mixture models vary between being parametric and non-parametric, where flexibility increases with $g$. From a theoretical perspective it is reasonable to choose Gaussian mixture models as the density functions coming from a mixture of Gaussians is dense in the set of all density functions with respect to the $L^1$-norm (\cite[Section 3.2]{MclachlanFiniteMixtures}). This flexibility and the provided parametrization make the mixture models a natural choice for the marginals. \newline

\subsection{A Blueprint of the Algorithm}\label{sec:PuttingItTogether}
The complete algorithm can be summarized as follows:

\begin{algorithm}[H]
\LinesNumbered
\label{algo:CompleteAlgorithm}
\SetAlgoLined
\KwIn{$X, \Sigma^0, \theta^0, g, \sigma_1, \ldots, \sigma_p, n_{max}, \epsilon_{converged}, M$}
\KwResult{$\Sigma, \theta$}
$n_{iter} \gets 0$\;
$\epsilon \gets \infty$\;
$\Sigma^t \gets \Sigma^0$\;
$\theta^t \gets \theta^0$\;
\While{$n_{iter} \leq {n_{max}}$ and $\epsilon > \epsilon_{converged}$}{
$\Sigma^{t+1} \gets$ solution of (\ref{eq:SolutionSigma})\;
\For{$x$ in $X$}{ 
Draw $M$ samples of $X | X_{\mathbf{obs}} = x_{\mathbf{obs}}$, under $\left(\theta^t, \Sigma^t\right)$\; \label{ln:RandomDraw}
}
$\theta^{t+1} \gets$ solution of (\ref{CopulaEMAlgoMonteCarloSum})\; 
$\label{ln:epsilon}\epsilon \gets \Vert \Sigma^{t+1} - \Sigma^{t} \Vert + \Vert \theta^{t+1} - \theta^{t} \Vert$\;
$\theta^{t} \gets \theta^{t+1}$\;
$\Sigma^{t} \gets \Sigma^{t+1}$\;
$n_{iter} \gets n_{iter} + 1$\;
}
\KwRet{$\Sigma^t, \theta^t$}
\caption{Blueprint for the EM algorithm for the Gaussian copula model}
\end{algorithm}
For the Monte Carlo EM Algorithm, \cite{WeiStochasticEM} propose to stabilize the parameters with a rather small $M$ and increase it substantially in the latter steps of the algorithm. This seems to be reasonable for line \ref{ln:RandomDraw} of Algorithm \ref{algo:CompleteAlgorithm} as well. \newline 
If we have no a priori knowledge about the marginals and we model them semiparametrically, we propose to choose $\theta^0$ such that the cumulative distribution functions of the mixture models fit to the ecdf of the observed data points. The number of components $g$ in the mixture model can then be chosen such that its corresponding $\theta^0$ provides a good approximation. For $\sigma_1, \ldots, \sigma_p$ we heuristically use the rule-of-thumb for the kernel density estimation by \cite{SilvermanKernelDensity}, where we replace the number of observations by $g$ and estimate the standard deviation using the observed data points only.

\section{Simulation Study}\label{sec:SimulationStudy}
In order to evaluate the method, we conduct a simulation study. In particular, we elaborate the issues arising from an MAR mechanism, leading to biased estimators for the marginals when deploying a two-step approach as \cite{GenestSemiparametricEstimation}. \newline 
We emphasize, that our data-generating process is fundamentally different from simulation studies based on an MCAR mechanism. In fact, with MCAR there is no need for an iterative approach like the EM algorithm. As the marginals can be estimated by the ecdfs of the observed data points consistently, we can use covariance estimators for MCAR (as \cite{LouniciCovarianceEstimation}) on
$$y_{ij} = \Phi^{-1}\left(\hat{F_j}(x_{ij})\right), i = 1, \ldots, N; j = 1, \ldots, p,$$
where $\hat{F_j}$ is the ecdf of the existing observations of column $j$. Alternatively, one can employ the method of \cite{WangCopulaPrecisionEstimation}.
\subsection{Setup}\label{sec:SimulationStudySetup}
We consider a two-dimensional data set\footnote{We would have liked to include the setup of the simulation study of \cite{SongEMCopula}. However, the missing mechanism can neither be extracted from the paper nor did the authors provide them on request.} with a priori unknown marginals $F_1$ and $F_2$, whose copula is a Gaussian copula with correlation parameter $\rho \in \mathbb[-1, 1]$. The marginals are chosen to be $\chi^2$ with $6$ and $7$ degrees of freedom. The data matrix $D \in \mathbb{R}^{N \times 2}$ keeps $N$ (complete) observations of the random vector. We enforce the following missing data mechanism:
\begin{enumerate}
\item Remove every entry in the data matrix $D$ with probability $0 \leq p_{MCAR} < 1$. The resulting data matrix (with missing entries) we denote as $D^{MCAR} = \left(D_{kj}^{MCAR}\right)_{k, j}$.
\item If $D_{k1}^{MCAR}$ and $D_{k2}^{MCAR}$ are observed, remove $D^{MCAR}_{k2}$ with probability 
\begin{equation*}
\begin{aligned}
\mathbb{P}\left(R_{2} = 0 | X_1 = D_{k1}, X_2 = D_{k2}\right) &= \mathbb{P}\left(R_{2} = 0 | X_1 = D_{k1}\right) \\
&= \left(1 + \exp\left(- \left(\beta_0 + \beta_1  \Phi^{-1}\left(F_1\left(D_{k1}\right)\right)\right)\right)\right)^{-1}.
\end{aligned}
\end{equation*}
We call the resulting data matrix $D^{MAR}$.
\end{enumerate}
The resulting data set is MAR. Besides $p_{MCAR}$, the parameters $\beta_0$ and $\beta_1$ control how many entries are absent in the final data set. Assuming that $\rho > 0$, $\beta_1 > 0$ and $|\beta_0|$ are not too large, the ecdf of the observed values of $X_2$ is shifted to the left compared to the underlying distribution function (changing the signs of $\beta_1$ and/or $\rho$ may change the direction of the shift, but the situation is analogous). We used this procedure to generate Figure \ref{fig:introduction_example} with $N = 200$, $\rho = 0.5$, $\beta_0 = 0$ and $\beta_1 = 2$. We observe that we could estimate the marginal distribution of $X_1$ using the ecdf of the observed data. 
\begin{figure}
\includegraphics[width=\textwidth]{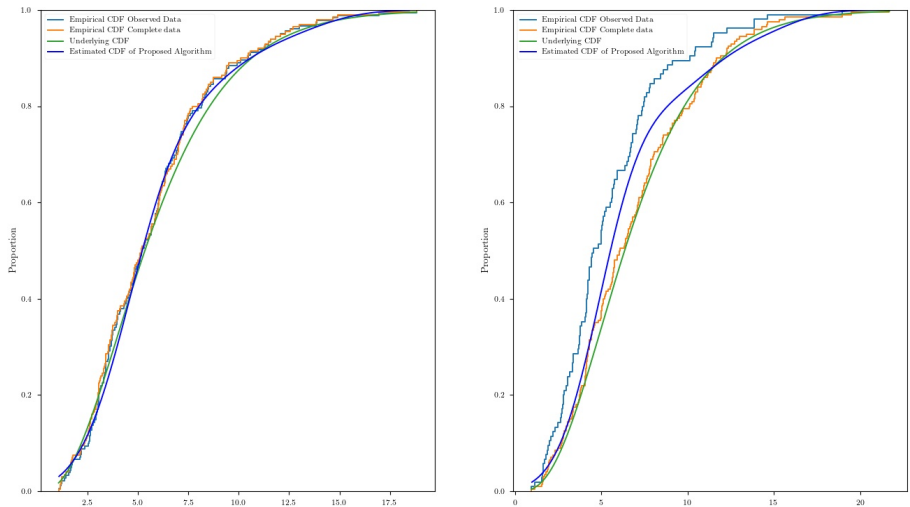}
\caption{Ecdf's of $X_1$ and $X_2$ using the data matrices $D^{MAR}$ (light blue line) and $D$ (orange line), the underlying cumulative distribution functions (green line) and the final estimates of the method of Section \ref{sec:EMAlgo} for the marginals (dark blue line).}
\label{fig:ExampleResult}
\end{figure}
\subsection{Adapting the EM Algorithm}
For the setup in Section \ref{sec:SimulationStudySetup} we choose $g=15$, for which we saw a sufficient flexibility for the marginal distributions. $\theta^0$ is then chosen by fitting the marginals to the existing observations. $\Sigma^0$ we set as the identity matrix. For $M$, we observed that with $M=20$, $\theta$ stabilizes after around $10$ steps. Cautiously, we run $20$ steps before we increase $M$ to $1000$ for which we run another 5 steps. We stop the algorithm, when the condition $\Vert\Sigma^{t+1} - \Sigma^{t}\Vert_1 < 10^{-5}$ is fulfilled.

\subsection{Results}
\begin{figure}[h]
\centering
\includegraphics[width=0.8\textwidth]{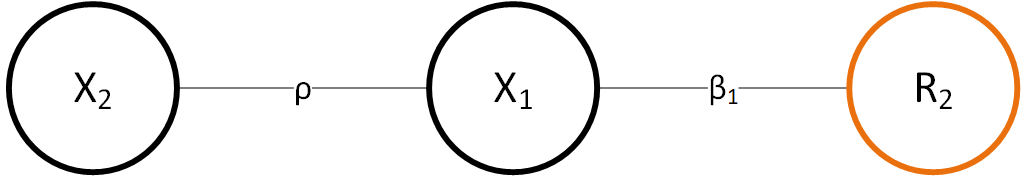}
\caption{Dependency graph for $X_1, X_2$ and $R_2$. $X_2$ is independent of $R_2$ if either $X_1$ and $X_2$ are independent ($\rho = 0$) or if $X_1$ and $R_2$ are independent ($\beta_1 = 0$).}
\label{fig:dependency_graph}
\end{figure}

\begin{figure}
\centering
\includegraphics[width=0.8\textwidth]{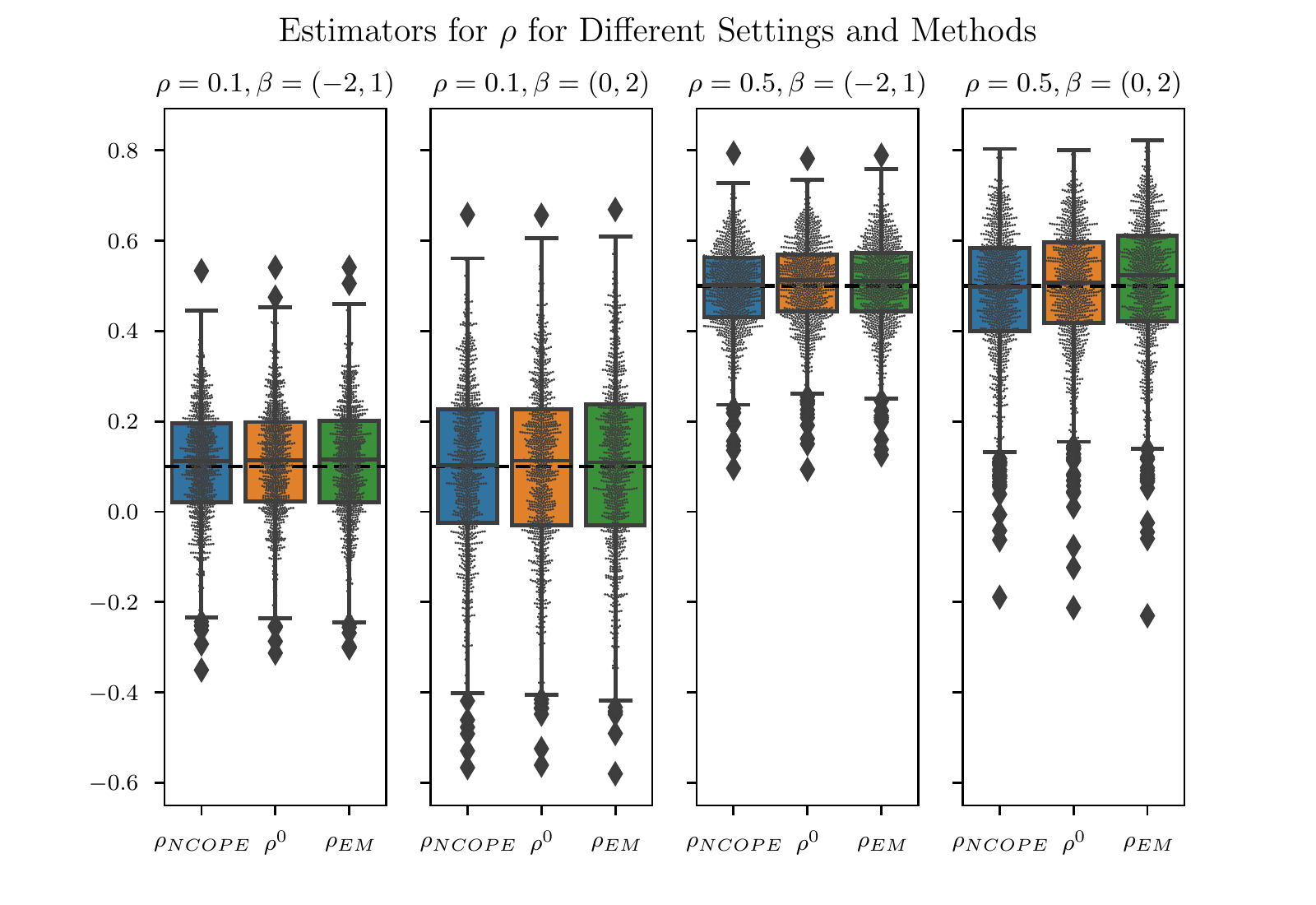}
\caption{Estimators for $\rho$ for different methods and settings. $\rho_{SCOPE}, \rho_{EM}, \rho^0$ are the estimators for $\rho$ of \cite{SongEMCopula}, Section \ref{sec:EMAlgo} of this paper and the gold standard, respectively. All methods deliver similar results in every setting. The estimation of the marginals does not have a great effect on the estimator for $\rho$, as the estimations of $\rho^0$ are not clearly superior.}
\label{fig:results_rho}
\end{figure}

\begin{figure}
\centering
\includegraphics[width=0.8\textwidth]{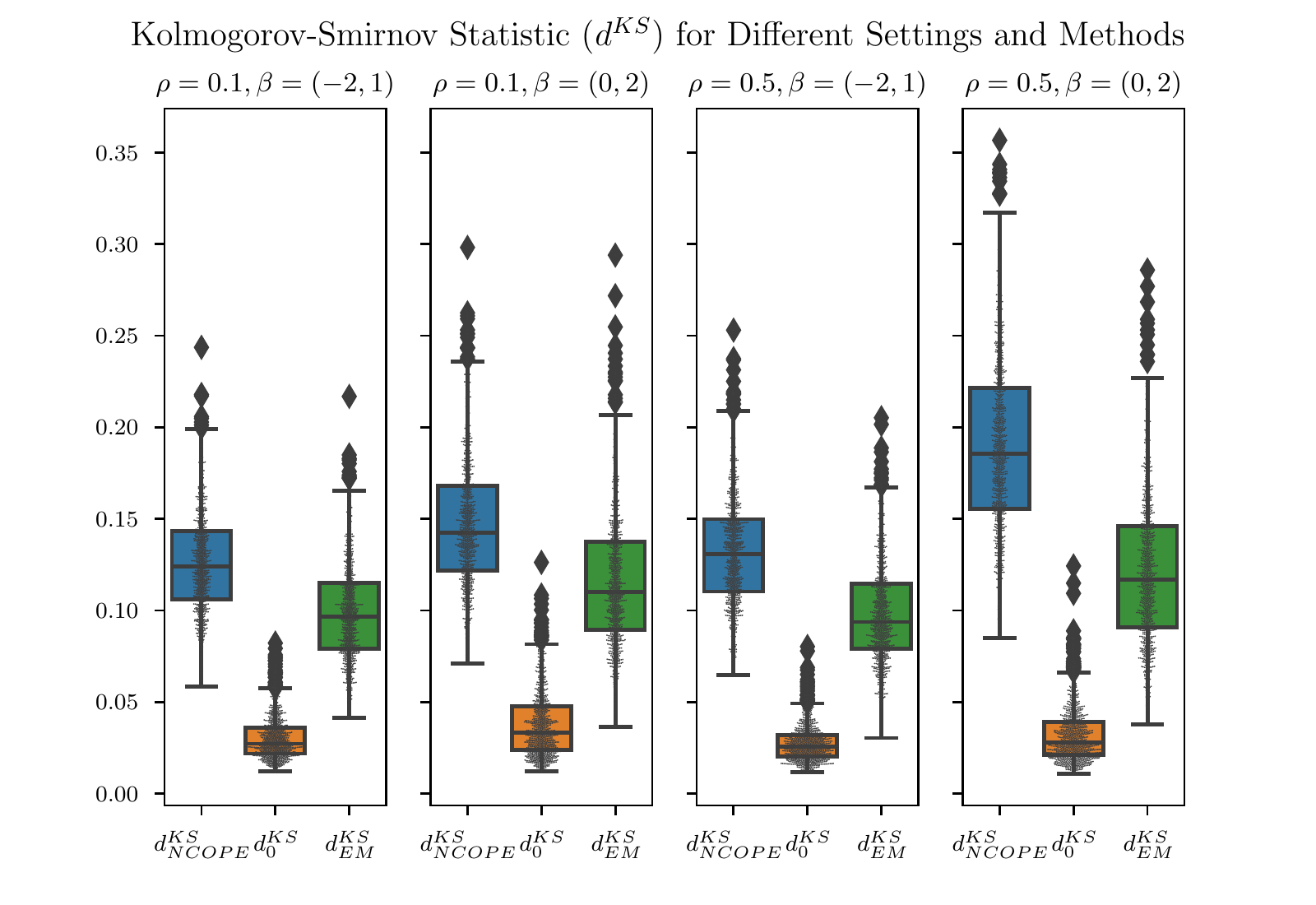}
\caption{Kolmogorov-Smirnov test statistic for different methods and settings. $d^{KS}_{SCOPE}, d^{KS}_{EM}, d^{KS}_0$ are the test statistics for \cite{SongEMCopula}, Section \ref{sec:EMAlgo} of this paper and the gold standard, respectively. The proposed method significantly improves the test statistic compared to \cite{SongEMCopula}. A higher correlation $\rho$ leads to a bigger difference in the test statistics between SCOPE and the proposed method. The estimation of the marginals is relevant, as $d_0^{KS}$ is substantially lower.}
\label{fig:results_kstest}
\end{figure}

We investigate four different settings of the setup in Section \ref{sec:SimulationStudySetup}. We vary the correlation $\rho \in \{0.1, 0.5\}$ and the missing mechanism parameters $\beta = (\beta_0, \beta_1) \in \{(-1, 1), (0, 2)\}$. As a competitor, we compare our method with the method of \cite{SongEMCopula}, where the marginal distributions are estimated by the observed data only. We call this algorithm the "Simplified COPula Estimator" or SCOPE for short. As a gold standard, we compute only the covariance structure  by applying an EM algorithm to the Gaussian observations $z_{lk} = \Phi^{-1}\left(F_k(x_{lk})\right), l = 1, \ldots, N, k = 1,2$. The idea is to eliminate the difficulty of finding estimators for the marginals.  \newline
For every one of those four settings we run 1000 simulations. To evaluate the methods, we look at two different aspects. \newline
First, we compare the estimators for $\rho$. The results are depicted in Figure \ref{fig:results_rho}. We see, that no method is clearly superior in estimating $\rho$. As even knowing the marginal distributions does not lead to substantially better estimators, we deduce that (at least in our setting) the quality of the estimators for the marginals is almost negligible for obtaining a good estimator for the copula.\newline
Second, we calculate a two-sample Kolmogorov-Smirnov (KS) test statistic as described in \cite[Section~5]{FasanoKSTest}. For that, we draw each time \SI{10000} data points from the learned joint distribution. We describe the details of the sampling procedure in the Appendix \ref{sec:DrawSamplesJointDistribution}. With those draws, we calculate the two-sample KS test statistic between the real distribution and the competitors. The results are depicted in Figure \ref{fig:results_kstest}. It shows, that the quality of the learned joint distribution depends highly on the estimation of the marginals, see \cite{thurowimputing} for a similar approach in the univariate setting. \newline 
Additionally, we see that the benefit of the proposed algorithm is larger in the case of high correlation. This is in line with the intuition that if the correlation is vanishing, the two random variables $X_1$ and $X_2$ are independent. Thus, the missingness probability $R_2$ and $X_2$ are independent. (Note that there is a difference from the case, where $\rho \neq 0$, and hence the missingness probability $R_2$ is \textbf{conditionally independent} from $X_2$ given $X_1$.) In that case, we can estimate the marginal of $X_2$ using the ecdf of the observed data points. Hence, for small $\rho$, SCOPE is almost consistent. An illustration can be found in Figure \ref{fig:dependency_graph}.
\newline 
\section{Discussion and Outlook}\label{sec:Discussion}
In this paper we have investigated the estimation of the Gaussian copula and the marginals with an incomplete data set. If the data is Missing At Random we have shown, that a consistent estimate of a marginal distribution depends on the copula and other marginals. Further, we derived a rigorous EM algorithm based on Monte Carlo integration that can still be applied. It works by iteratively finding the marginal distributions and the copula and is hence similar to known methods for complete data sets. \newline 
However, the EM algorithm relies on a complete parametrization of the marginals. In case there is no prior knowledge about them, we presented the novel idea to employ semiparametric mixture models. Although this is practically always a misspecification of the marginal distributions, our simulation study revealed that the combination of the EM algorithm and the semiparametric marginals delivers better estimates for the joint distribution than the currently used algorithms of \cite{SongEMCopula} and \cite{ZhaoUdellCopulaEM}. The intuition behind the algorithm is the following: \newline
Given a current state of knowledge about the copula $C^t$ and the marginals $F^t$, we can derive a probabilistic approximation of the location of the missing values. This is achieved by generating samples for the missing values given the observed values. With the information gained on the location of the data points and $C^t$, we can derive new estimators for the marginals $F^{t+1}$, leading to updated scores $u^{t+1} = F^{t+1}\left(x \right)$, which can then in turn be utilized to find an improved copula $C^{t+1}$. 
\newline
We note that the focus of the paper is on estimating the joint distribution without precise specification of its subsequent use. Therefore, we have not discussed imputation methods (see, e.g. \cite{rubin1996multiple}, \cite{van2018flexible}, \cite{ramosaj2019predicting}, \cite{ramosaj2020cautionary}) in more detail. However, some researchers have used the Gaussian copula model as a device for Multiple Imputation (MI) with some success (\cite{ZhaoUdellCopulaEM}, \cite{HollenbachWrongEM}, \cite{HouariWrongMarginals}), although all mentioned approaches do not see the importance of the modeling of the marginals. The resulting complete data sets can be used for inference. This seems odd, since we can derive all statistics from the learned joint distribution. However, it is not unusual to use a different model for imputation than for the analysis (\cite{SchaferMI}), partly because no fully parametric model for the analysis is used. In that case, the proposed procedure finds reasonable draws from the conditional distribution of the missing values given the observed data. The findings of Section \ref{sec:SimulationStudy} translate into better draws for the missing values.\newline
Besides that, the proposed procedure is interesting for applications focusing on marginal distributions. At the Battery Cell Competence Center, BMW builds battery cells on a prototype scale. A key performance indicator (KPI) is assigned to every battery cell at the end of the production line. However, some products do not reach the end of the production process because a decision system relying on observed measurements sorts out pieces. With the proposed method, one can assess the distribution of the KPI over all products and evaluate the performance of the decision system without intervening in the production process. As the data is confidential, the concrete application can not be shown here. \newline
With respect to future research, different aspects might be worth investigating:
\begin{itemize}
\item Maximize Equation (\ref{CopulaEMAlgoMonteCarloSum}) with respect to $\theta_1, \ldots, \theta_p$ sequentially. The derivative with respect to only one specific $\theta_l$ is faster to calculate and the involved terms demand less memory. It is again an application of the ECM algorithm of \cite{MengECM}.
\item Include the weights and bandwidths of the mixture models (Equation (\ref{eq:MixtureMarginalsSimplified})) to the parameters and examine other kernels like the Epanechnikov kernel.
\item Develop methods to select $g$ in Equation (\ref{eq:MixtureMarginalsSimplified}) (similar to \cite{MclachlanFindg} for complete data sets).
\item Generalize the approach to different parametric copulas $C_\psi$. In particular, when the copula $C_\psi(u_1, \ldots, u_p) = G_\psi(G^{-1}_1(u_1), \ldots, G^{-1}_p(u_p))$ and $G_\psi$ is the multivariate distribution function of an exponential family distribution with marginals $G_1, \ldots, G_p$, then the algorithm of Section \ref{sec:EMAlgo} also applies and there is a solution that depends on an expected statistic only.
\end{itemize}
Given the numerous opportunities for future research and the promising results of our method, we are looking forward to more interesting contributions in the field of semiparametric density estimation in the case of missing data.
\begin{appendices}
\section{Appendix}\label{sec:Appendix}
\subsection{Proof of Conditional Distribution}
\label{sec:ProofProposition}
\begin{proof}[Proof of Proposition  \ref{ConditionalDistributionGaussianCopula}]
We prove in the order of the Proposition, which is a multivariate generalization of \cite{kaarik}.
\begin{enumerate}
\item Inspect the conditional density function:
\begin{equation*}
\begin{aligned}
f(x_\mathbf{T} | X_\mathbf{S} = x_\mathbf{S}) &= 
\frac{|\Sigma|^{-\frac{1}{2}} \exp\left(-\frac{1}{2} z^T \left(\Sigma^{-1} - I\right)  z\right) \prod_{j=1}^p f_j(x_j)}{|\Sigma_{\mathbf{S}, \mathbf{S}}|^{-\frac{1}{2}} \exp\left(-\frac{1}{2} z_\mathbf{S}^T \left(\Sigma_{\mathbf{S}, \mathbf{S}}^{-1} - I\right)  z_\mathbf{S}\right) \prod_{j \in \mathbf{S}} f_j(x_j)} \\ 
&= \frac{|\Sigma|^{-\frac{1}{2}} \exp\left(-\frac{1}{2} z^T \Sigma^{-1}  z\right) \exp(\frac{1}{2} z^T z) \prod_{j=1}^p f_j(x_j)}
{|\Sigma_{\mathbf{S}, \mathbf{S}}|^{-\frac{1}{2}} \exp\left(-\frac{1}{2} z_\mathbf{S}^T \Sigma_{\mathbf{S}, \mathbf{S}}^{-1}  z_\mathbf{S}\right) \exp(\frac{1}{2} z_{\mathbf{S}}^T z_{\mathbf{S}})\prod_{j \in \mathbf{S}} f_j(x_j)} \\
&= \frac{|\Sigma|^{-\frac{1}{2}} \exp\left(-\frac{1}{2} z^T \Sigma^{-1}  z\right)  \exp(\frac{1}{2} z_{\mathbf{T}}^T z_{\mathbf{T}}) \prod_{j \in \mathbf{T}} f_j(x_j)}
{|\Sigma_{\mathbf{S}, \mathbf{S}}|^{-\frac{1}{2}} \exp\left(-\frac{1}{2} z_\mathbf{S}^T \Sigma_{\mathbf{S}, \mathbf{S}}^{-1}  z_\mathbf{S}\right)}
\end{aligned}
\end{equation*}
Using well-known factorization lemmas using the Schur complement (see for example \cite[Section 4.3.4]{MurphyML}) applied on $\Sigma^{-1}$ we encounter
\begin{equation}
\label{eq:ConditionalDistributionCopula}
f(x_\mathbf{T} | X_\mathbf{S} = x_\mathbf{S}) = |\Sigma'|^{-\frac{1}{2}} \exp\left(-\frac{1}{2} (z_\mathbf{T} - \mu)^T \Sigma'^{-1}  (z_\mathbf{T} - \mu)\right) \exp\left(\frac{1}{2} z_{\mathbf{T}}^T z_{\mathbf{T}}\right) \prod_{j \in \mathbf{T}} f_j(x_j).
\end{equation}
\item The distribution of $$\Phi^{-1}\left(F(X_\mathbf{T})\right) | X_\mathbf{S} = x_\mathbf{s}$$ follows with a change of variable argument. Using Equation (\ref{eq:ConditionalDistributionCopula}), we observe for any measurable set $A$
\begin{equation*}
\begin{aligned}
&\mathbb{P}\left(\left(\Phi^{-1}\left(F(X_\mathbf{T})\right) | X_\mathbf{S} = x_\mathbf{s}\right) \in A\right) \\
&= \int_{F^{-1}\left(\Phi\left(A\right)\right)}  |\Sigma'|^{-\frac{1}{2}} \exp\left(-\frac{1}{2} (z_\mathbf{T} - \mu)^T \Sigma'^{-1}  (z_\mathbf{T} - \mu)\right) \exp\left(\frac{1}{2} z_{\mathbf{T}}^T z_{\mathbf{T}}\right) \prod_{j \in \mathbf{T}} f_j(x_j) dx_{\mathbf{T}} \\
&= \int_{A} \phi_{\mu, \Sigma'}(q_\mathbf{T}) dq_\mathbf{T},
\end{aligned}
\end{equation*}
where we used in the second equation the transformation $q_\mathbf{T} = \Phi^{-1}\left(F(x_\mathbf{T})\right)$ and the fact that 
$$\left\vert D\left(\phi^{-1}\left(F\left(x_\mathbf{T}\right)\right)\right)\right\vert = 2 \pi ^{\frac{|\mathbf{T}|}{2}}\exp\left(\frac{1}{2}\left(\Phi^{-1}\left(F(x_\mathbf{T})\right)\right)^T\left(\Phi^{-1}\left(F(x_\mathbf{T})\right)\right)\right) \prod_{j \in \mathbf{T}} f_j(x_j).$$
\item This proof is analogous to the one above and we finally obtain
\begin{align*}
\int h(x_{\mathbf{T}}) f(x_\mathbf{T} | X_\mathbf{S} = x_\mathbf{S}) dx_\mathbf{T}= \int h\left(F^{-1}\left(\Phi\left(z_{\mathbf{T}}\right)\right)\right) \phi_{\mu, \Sigma'}(z_{\mathbf{T}})dz_{\mathbf{T}}.
\end{align*}
\end{enumerate}
\end{proof}
\subsection{Closed-form Solution of E-Step for $\theta = \theta^t$}
\begin{theorem}\label{ConditionalExpectationGaussianizedEM}
Assume w.l.o.g., that $x = (x_{\mathbf{obs}}, x_\mathbf{{mis}})$ and let $\left(z_{\mathbf{obs}, \theta^t}, z_{\mathbf{mis}, \theta^t}\right) = z_{\theta^t} := \Phi^{-1}\left(F_{\theta^t}(x)\right)$. Then it holds, that
$$\mathbb{E}_{\Sigma^t, \theta^t}\left(-\frac{1}{2}\ln(|\Sigma|) - \frac{1}{2}{z_{\theta^t}}^T \Sigma^{-1} z_{\theta^t} | X_{\mathbf{obs}} = x_{\mathbf{obs}}\right) = -\frac{1}{2}\ln(|\Sigma|) - \frac{1}{2}tr\left(\Sigma^{-1}V\right),$$
where $V =  \begin{pmatrix}
  z_{\mathbf{obs}, \theta^t} {z_{\mathbf{obs}, \theta^t}}^T & z_{\mathbf{obs}, \theta^t} \mu^T \\
  \mu {z_{\mathbf{obs}, \theta^t}}^T & \Sigma' + \mu \mu^T
  \end{pmatrix}$, $\Sigma' = \Sigma^t_{\mathbf{mis}, \mathbf{mis}} - \Sigma^t_{\mathbf{mis}, \mathbf{obs}} {\Sigma^t_{\mathbf{obs}, \mathbf{obs}}}^{-1}\Sigma^t_{\mathbf{obs}, \mathbf{mis}}$, $\mu = \Sigma^t_{\mathbf{mis}, \mathbf{obs}}{\Sigma^t_{\mathbf{obs}, \mathbf{obs}}}^{-1}z_{\mathbf{obs}, \theta^t}$ and $z_{\mathbf{obs}, \theta^t} = \Phi^{-1}\left(F_\textbf{obs}^{\theta^t}(x_{\textbf{obs}})\right)$. 

\end{theorem}
\begin{proof}
\begin{equation*}
\begin{split}
&\mathbb{E}_{\Sigma^t, \theta^t}\left(-\frac{1}{2}\ln(|\Sigma|) - \frac{1}{2}z_{\theta^t}^T \Sigma^{-1} z_{\theta^t} | X_{\mathbf{obs}} = x_{\mathbf{obs}}\right) \\
&=\mathbb{E}_{\Sigma^t, \theta^t}\left(-\frac{1}{2}\ln(|\Sigma|) - \frac{1}{2}{\left(\Phi^{-1}\left(F_{\theta^t}(x)\right)\right)}^T \Sigma^{-1} \left(\Phi^{-1}\left(F_{\theta^t}(x)\right)\right) | X_{\mathbf{obs}} = x_{\mathbf{obs}}\right) \\
&=-\frac{1}{2}\ln(|\Sigma|) -  \frac{1}{2} \int {\left(\Phi^{-1}\left(F_{\theta^t}(x)\right)\right)}^T \Sigma^{-1} \left(\Phi^{-1}\left(F_{\theta^t}(x)\right)\right)f_{\theta^t, \Sigma^t}\left(x_{\mathbf{mis}} | X_{\mathbf{obs}} = x_{\mathbf{obs}}\right) dx_{\mathbf{mis}} 
\end{split}
\end{equation*}
We are now applying Proposition \ref{ConditionalDistributionGaussianCopula}. We then encounter
\begin{equation*}
\begin{split}
&-\frac{1}{2}\ln(|\Sigma|) -  \frac{1}{2}\int {\left(\Phi^{-1}\left(F_{\theta^t}(x)\right)\right)}^T \Sigma^{-1} \Phi^{-1}\left(F_{\theta^t}(x)\right) f_{\theta^t, \Sigma^t}\left(x_{\mathbf{mis}} | X_{\mathbf{obs}} = x_{\mathbf{obs}}\right) dx_{\mathbf{mis}} \\
&= -\frac{1}{2}\ln(|\Sigma|) - \frac{1}{2} \int  z_{\theta^t}^T \Sigma^{-1} z_{\theta^t} \phi_{\Sigma', \mu}(z_{\mathbf{mis}, {\theta^t}}) dz_{\mathbf{mis}, {\theta^t}} \\
& = -\frac{1}{2}\ln(|\Sigma|) - \frac{1}{2} \int tr(z_{\theta^t} z_{\theta^t}^T \Sigma^{-1}) \phi_{\Sigma', \mu}(z_{\mathbf{mis}, {\theta^t}}) dz_{\mathbf{mis}, {\theta^t}} \\
& = -\frac{1}{2}\ln(|\Sigma|) - \frac{1}{2} tr\left(\Sigma^{-1}  \int z_{\theta^t} z_{\theta^t}^T \phi_{\Sigma', \mu}(z_{\mathbf{mis}, {\theta^t}}) dz_{\mathbf{mis}, {\theta^t}}\right).
\end{split}
\end{equation*}
The last integral is understood elementwise. Taking a closer look at the integral, we see
\begin{equation*}
\begin{split}
\int z_{\theta^t} z_{\theta^t}^T \phi_{\Sigma', \mu}(z_{\mathbf{mis}, {\theta^t}}) dz_{\mathbf{mis}, {\theta^t}} &= \int \left(z_{\mathbf{obs}, {\theta^t}}, z_{\mathbf{mis}, {\theta^t}}\right) \left(z_{\mathbf{obs}, {\theta^t}}, z_{\mathbf{mis}, {\theta^t}}\right)^T \phi_{\Sigma', \mu}(z_{\mathbf{mis}, {\theta^t}}) dz_{\mathbf{mis}, {\theta^t}} \\
&=  \begin{pmatrix}
  z_{\mathbf{obs}, {\theta^t}} z_{\mathbf{obs}, {\theta^t}}^T & z_{\mathbf{obs}, {\theta^t}} \mu^T \\
  \mu z_{\mathbf{obs}, {\theta^t}}^T & \Sigma' + \mu \mu^T
  \end{pmatrix}
.\end{split}
\end{equation*} 
\end{proof}
\subsection{Maximizer of $\argmax_{\Sigma, \Sigma_{jj} = 1 \forall j = 1, \ldots, p} \lambda(\theta^t, \Sigma | \theta^t, \Sigma^t)$}
\label{sec:appendix_max_sigma}
We are interested in 
$$\argmax_{\Sigma_{jj} = 1 \forall j = 1, \ldots, p} l(\Sigma) := \argmax_{\Sigma_{jj} = 1 \forall j = 1, \ldots, p} -\log \left(|\Sigma|\right) - tr\left(\Sigma^{-1} S\right),$$
where $\Sigma, S \in \mathbb{R}^{p \times p}$ are positive definite matrices. 
Clearly,
$$\Sigma_{jj} = 1 \iff 1 = e_j^{T}  \Sigma e_j = tr\left(e_j^{T}  \Sigma e_j\right)= tr\left(e_j e_j^{T} \Sigma\right).$$
Hence, using the Lagrangian, we obtain the following function to optimize
$$L(\Sigma, \lambda) = - \log \left(|\Sigma|\right) - tr\left(\Sigma^{-1} S\right) + \sum_{j=1}^{p} \lambda_j \left( tr\left(e_j e_j^{T} \Sigma\right) - 1\right).$$
Using the identities $\frac{\partial tr(AX)}{\partial X} = A$, $\frac{\partial tr(AX^{-1})}{\partial X} = -X^{-1}AX^{-1}$, $\frac{\partial \log(|X|)}{\partial X} = X^{-1}$, we obtain the derivative with respect to $\Omega$
$$\frac{\partial L}{\partial \Sigma} = -\Sigma^{-1} + \Sigma^{-1} S \Sigma^{-1} - \left(\sum_{j=1}^{p} \lambda_j \left( e_j e_j^{T} \right) \right)  \overset{!}{=} 0.$$
This is equivalent to 
$$-\Omega +\Omega S \Omega   = D_\lambda,$$
where $D_\lambda$ is the diagonal matrix with entries $\lambda = \left(\lambda_1, \ldots, \lambda_p\right)$ and $\Omega := \Sigma^{-1}$. We see that the scaling of $S$ by $a \in \mathbb{R}_{>0}$ leads in general to a different solution $\Omega$ and hence the estimator is not invariant under strictly monotone linear transformations of $S$.
\noindent We can also formulate the task as the convex optimization problem 
$$\argmin_{\left(\Omega^{-1}\right)_{ii} = 1 \forall i = 1, \ldots, p} -\log \left(|\Omega|\right) + tr\left(\Omega S\right).$$
\subsection{Drawing Samples of the Joint Distributions}
\label{sec:DrawSamplesJointDistribution}
\subsubsection{Estimators of the Percentile Function}
\begin{itemize}
\item In case of SCOPE, consider the observed data points, which we assume to be ordered $y_1 \leq \ldots \leq y_N$. We use the following linearly interpolated estimator for the percentile function:
\[
    \widehat{F^{-1}}(u)  = \left\{\begin{array}{lr}
        y_1 & \text{for } u \leq \frac{1}{N+1}\\
        y_N, & \text{for } u > \frac{N}{N+1}\\
        \frac{u - \frac{i}{N + 1}}{\frac{i + 1}{N + 1} - \frac{i}{N + 1}}(y_{i + 1} - y_{i}) + y_{i}, & \text{for } u\in \bigg(\frac{i}{N + 1}, \frac{i + 1}{N +1}\bigg]
        \end{array}\right\}
  \]
\item To estimate the percentile function for the mixture models, we choose with equal probability (all Gaussians have equal weight) one component of the mixture and then draw a random number with its mean $\theta_{ij}$ and standard deviation $\sigma_i$, $i=1, \ldots, p, \text{ } j = 1, \ldots, g$. In this manner, we generate $N'$ samples $y'_1, \ldots, y'_{N'}$. The estimator for the percentile function is then chosen to be analogous to the one above. A higher $N'$ leads to a more exact result. We choose $N'$ to be \SI{10000}{}.
\end{itemize}
\subsubsection{Sampling}
Given an estimator $\widehat{\rho}$ and estimators for the percentile functions $\widehat{F_1^{-1}}, \widehat{F_2^{-1}}$, we obtain samples from the learned joint distribution with 
$$y_l = \left(y_{l1}, y_{l2}\right) = \left(\widehat{F_1^{-1}}(u_{l1}), \widehat{F_2^{-1}}(u_{l2})\right) = \left(\widehat{F_1^{-1}}\left(\Phi\left(z_{l1}\right)\right), \widehat{F_2^{-1}}\left(\Phi\left(z_{l2}\right)\right)\right), l = 1, \ldots, K,$$ where $z_l = \left(z_{l1}, z_{l2}\right), l = 1, \ldots, K$ are draws from a bivariate normal distribution with mean $0$ and covariance $ \begin{pmatrix}
  1 & \widehat{\rho} \\
  \widehat{\rho} & 1
  \end{pmatrix}
$. 
In the case of the gold standard, we set $\widehat{F_j^{-1}} = F_j^{-1}, j = 1,2$. We obtain samples of the real underlying distribution by using the correct percentile functions as in the gold standard and additionally $\widehat{\rho} = \rho$.
\end{appendices}

\bibliographystyle{unsrt}
\bibliography{bibliography}

\end{document}